\documentclass[12pt]{article}
 \usepackage{amsmath}

\usepackage{fancyhdr}
\usepackage{graphicx}
\usepackage{color}
\usepackage{graphicx} 
\usepackage{amsthm}
\usepackage{amsmath}
\usepackage{amsfonts}
\usepackage{amssymb}
\usepackage{algorithm, algpseudocode}
\usepackage{xcolor}
\usepackage{textcomp}
\usepackage{booktabs}
\usepackage{subcaption}
\usepackage[left=2.5cm,right=2.5cm,top=2.5cm,bottom=2.5cm]{geometry}
\begin{document}
\renewcommand{\headrulewidth}{0pt}
\fancyfoot[L]{ 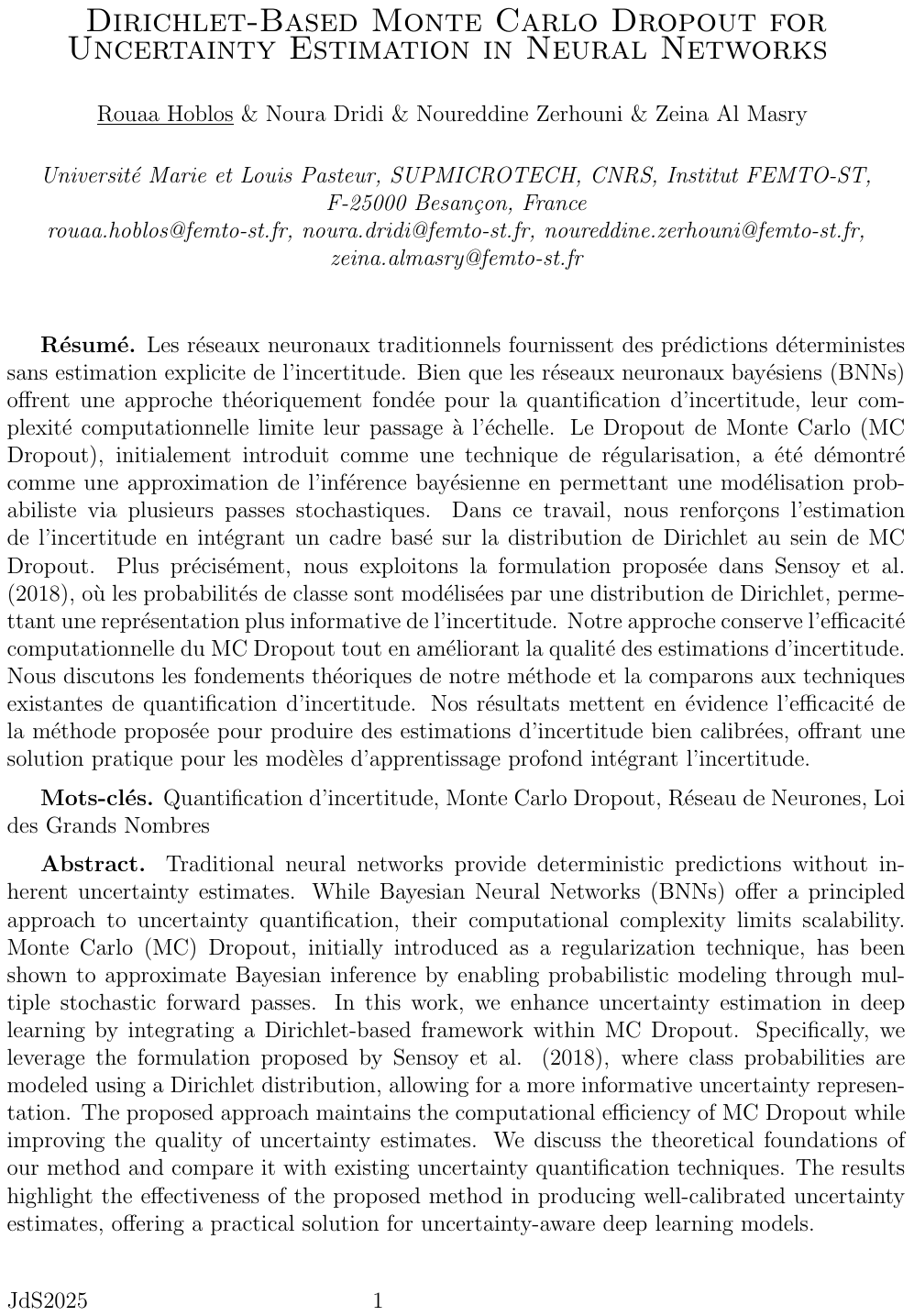}
\fancyhead[R]{ }


\begin{center}
{\Large
	{\sc Dirichlet-Based Monte Carlo Dropout for Uncertainty Estimation in Neural Networks}
}
\bigskip

\underline{Rouaa Hoblos} \& {Noura Dridi}  \& {Noureddine Zerhouni}  \& {Zeina Al Masry} 
\bigskip

{\it
{Université Marie et Louis Pasteur, SUPMICROTECH, CNRS, Institut FEMTO-ST, F-25000 Besançon, France} \\

rouaa.hoblos@femto-st.fr,
noura.dridi@femto-st.fr,
noureddine.zerhouni@femto-st.fr, 
zeina.almasry@femto-st.fr\\

}
\end{center}
\bigskip





{\bf Résumé.}  
Les réseaux neuronaux traditionnels fournissent des prédictions déterministes sans estimation explicite de l'incertitude. Bien que les réseaux neuronaux bayésiens (BNNs) offrent une approche théoriquement fondée pour la quantification d'incertitude, leur complexité computationnelle limite leur passage à l’échelle. Le Dropout de Monte Carlo (MC Dropout), initialement introduit comme une technique de régularisation, a été démontré comme une approximation de l'inférence bayésienne en permettant une modélisation probabiliste via plusieurs passes stochastiques.  Dans ce travail, nous renforçons l'estimation de l'incertitude en intégrant un cadre basé sur la distribution de Dirichlet au sein de MC Dropout. Plus précisément, nous exploitons la formulation proposée dans Sensoy et al. (2018), où les probabilités de classe sont modélisées par une distribution de Dirichlet, permettant une représentation plus informative de l'incertitude. Notre approche conserve l'efficacité computationnelle du MC Dropout tout en améliorant la qualité des estimations d'incertitude. Nous discutons les fondements théoriques de notre méthode et la comparons aux techniques existantes de quantification d'incertitude.
Nos résultats mettent en évidence l'efficacité de la méthode proposée pour produire des estimations d'incertitude bien calibrées, offrant une solution pratique pour les modèles d'apprentissage profond intégrant l'incertitude.  

{\bf Mots-clés.} Quantification d'incertitude, Monte Carlo Dropout, Réseau de Neurones, Loi des Grands Nombres

{\bf Abstract.} 
Traditional  neural  networks  provide deterministic  predictions  without  inherent  uncertainty  estimates.   While  Bayesian  Neural  Networks  (BNNs)  offer  a  principled approach  to  uncertainty  quantification,  their  computational  complexity  limits  scalability. Monte  Carlo  (MC)  Dropout,  initially  introduced  as  a  regularization  technique,  has  been shown to approximate Bayesian inference by enabling probabilistic modeling through multiple  stochastic  forward  passes.   In  this  work,  we  enhance  uncertainty  estimation  in  deep learning  by  integrating  a  Dirichlet-based  framework  within  MC  Dropout.   Specifically,  we leverage  the  formulation  proposed  by  Sensoy  et  al.  (2018),  where  class  probabilities  are modeled using a Dirichlet distribution, allowing for a more informative uncertainty representation.  The proposed approach maintains the computational efficiency of MC Dropout while improving  the  quality  of  uncertainty  estimates.   We  discuss  the  theoretical  foundations  of our method and compare it with existing uncertainty quantification techniques. 
The results highlight the effectiveness of the proposed method in producing well-calibrated uncertainty estimates, offering a practical solution for uncertainty-aware deep learning models.



\section{Introduction}
Effective decision-making relies on the ability to make precise and reliable predictions including quantifying uncertainty associated with these predictions, this means the confidence level to attribute to these predictions. This is particularly relevant for critical domains such as medical field or autonomous vehicle control. Several approaches have been developed, including Bayesian Neural Networks (BNNs) \cite{Neal95}, ensemble methods \cite{lakshminarayanan17}, and single deterministic models \cite{malinin18}.
BNNs estimate uncertainty by modeling neural network weights as probability distributions.
A variational interpretation of the BNN is proposed in \cite{Gal16}. The main result is that optimizing the loss function
of a NN with dropout is equivalent to a Bayesian variational approximation of a Gaussian process. This method enables probabilistic modeling by leveraging dropout at both training and inference stages, allowing multiple stochastic forward passes to estimate uncertainty. The method is computationally efficient compared to classical BNNs \cite{Neal95}, which are computationally expensive for posterior inference. On the other hand, ensemble methods train multiple models and quantify uncertainty using the variability among their predictions. Single deterministic methods, such as evidential deep learning \cite{sensoy18},
use a Dirichlet distribution to represent the class probability. The parameters of this distribution are related to the prediction and uncertainty. In \cite{sensoy18}, the authors propose a new method called Deep Evidential Classification (DEC), an evidential neural network to learn these parameters in classification tasks. In \cite{tsiligkaridis21}, the author proposes an Information Aware Dirichlet (IAD) network to learn a Dirichlet prior distribution on predictive distributions. In this work, we propose to use the same hypothesis for the class probability, however, we propose a new method to calculate Dirichlet parameters and therefore the uncertainty.
This method leverages MC Dropout while maintaining computational efficiency and improving uncertainty estimation through a theoretically grounded probabilistic formulation. The performance of the method is confirmed on tabular and image datasets and compared with the approach proposed in \cite{sensoy18} and \cite{tsiligkaridis21} using different metrics.

\section{Proposed approach}

Let $\{(X_i, Y_i)\}_{i=1}^{N}$ be a set of data samples, where $X_i$ represents an input (e.g., an image or a feature vector), and $Y_i$ is the corresponding label in a classification task, with $Y_i \in \mathcal{Y} = \{1,2,\dots,K\}$. A neural network parameterized by $\theta$ is used to predict the class probability distribution $ p_{\theta}(X_i) = (p_1, p_2, \dots, p_K)$, where $p_k$ is the predicted probability for class $k$, obtained by applying the softmax function to the network’s logits.

\noindent
\textbf{Uncertainty and Evidence.} 
 In \cite{sensoy18}, Sensoy has proposed calculating the uncertainty using evidence learned by the neural network. The NN outputs evidence which is used in the calculation of the parameters of a Dirichlet distribution. These parameters are used to calculate the predicted probabilities and the uncertainties associated to them. The uncertainty is given by:
\begin{equation}\label{eq:U}
    u=\frac{K}{S}
\end{equation}
with $S=\sum_{k=0}^K(e_k+1)$, with $e_k$  the evidence derived for the $kth$ singleton.
The authors \cite{sensoy18} propose to replace the softmax activation function, classically used for classification, by the \textit{ReLU} function to obtain positive values corresponding to the evidence. In \cite{tsiligkaridis21}, Tsiligkaridis uses the softplus activation function. The predicted probabilities are the expected value calculated using the properties of a Dirichlet distribution: $p_k = \frac{\alpha_k}{S}$  where $p_k$ is the probability associated to the $k^{th}$ class.
 
\noindent
\textbf{Dirichlet Based Uncertainty Estimation-MC Dropout (DBUE-Dropout).} As proposed in \cite{sensoy18}, the class probability vector $\mathbf{p} = (p_1, p_2, \dots, p_K)$ is assumed to follow a \textit{Dirichlet distribution} parameterized by $\boldsymbol{\alpha} = (\alpha_1, \alpha_2, \dots, \alpha_K)$:
\begin{equation}
    p(\mathbf{p} | \boldsymbol{\alpha}) = \frac{1}{B(\boldsymbol{\alpha})} \prod_{k=1}^{K} p_k^{\alpha_k - 1}
\end{equation}
where $B(\boldsymbol{\alpha})$ is the multivariate Beta function. 

\noindent
The variance of the Dirichlet distribution can be written as \cite{minka2000}:
\begin{equation}
    var(p_k) = \frac{\mathbb{E}[p_k](1-\mathbb{E}[p_k])}{1+\sum_{k}\alpha_k}.
\end{equation}

\noindent
Therefore,
 $S$ would be given by:
\begin{equation} \label{S}
    S = \sum_{k}\alpha_k = \frac{\mathbb{E}[p_k](1-\mathbb{E}[p_k])}{\text{Var}(p_k)} - 1
\end{equation}
where $\mathbb{E}[p_k]$ is the expected probability for class $k$ and $\text{Var}(p_k)$ is the variance associated to this predicted probability.

\noindent

In the context of classification, a NN architecture is designed to output the class probabilities. Besides, by activating the dropout on the test, we obtain a different probability for each new forward.
It is worth pointing out that MC dropout computes the standard deviation to estimate uncertainty; however, the standard deviation measures the variation but doesn't convey the full characteristics of a distribution. We propose to represent the probability of classes by the Dirichlet distribution. Indeed, a classification problem with K labels is a multinomial opinion equivalent to a Dirichlet probability distribution function\cite{jsang18}.
Given an input $X_i$, we perform $L$ stochastic forward passes with dropout enabled, obtaining $L$ probability vectors: $ p^{(1)}, p^{(2)}, \dots, p^{(L)}$. For each class $k$, the expected probability and variance are computed empirically.
Besides, we estimate $\mathbb{E}[p_k]$ and $\text{Var}(p_k)$ using the Law of Large Numbers (LLN) \cite{loeve1977elementary}. The LLN states that the empirical mean of a sequence of independent and identically distributed (i.i.d.) random variables converges to the true expectation as the number of samples increases. Despite its simplicity, this process makes our proposed method a fully coherent Bayesian method with a connection to evidence. 
Using the empirical estimates, $S$ is computed using \eqref{S} and used to obtain the final uncertainty measure as in \eqref{eq:U}. A higher S means more evidence and less uncertainty and vice versa. It is important to note that the uncertainty is calculated using the absolute value of $S$ to ensure it is non-negative.\\
%

\noindent
\textbf{Algorithm.} An algorithm of the proposed methodology is shown below (Algorithm 1). 
\begin{algorithm}
\caption{Dirichlet Based Uncertainty Estimation-MC Dropout}\label{alg:uncertainty}
\begin{algorithmic}
\Require Dataset $\{(X_i, Y_i)\}_{i=1}^{n}$, trained neural network $p_{\theta}(X)$, number of MC Dropout iterations $L$, number of classes $K$
\Ensure Uncertainty estimates $u_i$ for each sample $X_i$
\State \textbf{Initialize:} Neural network $p_{\theta}(X)$ with MC Dropout enabled
\For{each sample $X_i$ in the test set}
    \State Collect $L$ stochastic forward passes:
    \Statex \hspace{1cm} $p_i^{(l)} = p_{\theta}(X_i)$ for $l = 1, \dots, L$
    \State Determine the predicted class:
    \Statex \hspace{1cm} $c_i = \arg\max_k \left( \frac{1}{L} \sum_{l=1}^{L} p_i^{(l)}(k) \right)$
    \State Extract probabilities of the predicted class:
    \Statex \hspace{1cm} $p_i^{(l)}(c_i)$ for $l = 1, \dots, L$
    \State Compute the empirical expectation for predicted class:
    \Statex \hspace{1cm} $\mathbb{E}[p_i(c_i)] = \frac{1}{L} \sum_{l=1}^{L} p_i^{(l)}(c_i)$
    \State Compute the empirical variance:
    \Statex \hspace{1cm} $\text{Var}(p_i(c_i)) = \frac{1}{L-1} \sum_{l=1}^{L} \left( p_i^{(l)}(c_i) - \mathbb{E}[p_i(c_i)] \right)^2$
    \State Compute Dirichlet strength parameter for predicted class:
    \Statex \hspace{1cm} $S_i = \frac{\mathbb{E}[p_i(c_i)](1 - \mathbb{E}[p_i(c_i)])}{\text{Var}(p_i(c_i))} - 1$
    \State Compute uncertainty:
    \Statex \hspace{1cm} $u_i = \frac{K}{S_i}$
\EndFor
\State \textbf{Return:} Uncertainty estimates $\{u_i\}_{i=1}^{n}$
\end{algorithmic}
\end{algorithm}

\section{Results}

The proposed method is evaluated on both tabular and image datasets, specifically MNIST and Titanic. To assess the effectiveness of the UQ method in detecting data shifts, we introduce noisy data and Out-of-Distribution (OOD) datasets. For MNIST, Gaussian noise with a mean of 2 and standard deviation of 100 is added to test images, while for Titanic, noise is selectively applied to different features with a mean of either 0 or 1 and a maximum standard deviation of 3. The Fashion-MNIST (FMNIST) dataset serves as an OOD dataset for MNIST, whereas the Forest Fires dataset is used for Titanic. Training is conducted with 350 epochs and 100 Monte Carlo dropout samples ($p_{drop} = 0.25$) for Titanic, while MNIST is trained for 150 epochs with 100 dropout samples ($p_{drop} = 0.4$). Furthermore the method is compared to  DEC \cite{sensoy18} and IAD \cite{tsiligkaridis21}. For the DEC, the NN outputs the evidence calculated using the ReLU activation function, and the uncertainty is calculated using \eqref{eq:U}. As for the IAD, the NN outputs the evidence calculated using the Softplus activation function, and the uncertainty is calculated using  \eqref{eq:U}.

\begin{table}[h!]
    \centering
    \caption{Performance Metrics Across Titanic Dataset with Different Methods.}
    \resizebox{\textwidth}{!}{%
    \begin{tabular}{l|cccc|cccc}
        \toprule
        Method & \multicolumn{4}{c|}{Accuracy} & \multicolumn{4}{c}{Uncertainty}  \\
         & Train & Test & Noisy & OOD & Train & Test & Noisy & OOD \\
        \midrule
        IAD \cite{tsiligkaridis21} & 86.82 & 76.92  & \textbf{72.032} & \textbf{0.00} & 0.53 & 0.54 & 0.51 & 0.012 \\
        DEC \cite{sensoy18} & 88.22 & \textbf{80.42} & 65.73 & \textbf{0.00} & 0.59 & 0.57 & 0.48 & 0.0033  \\
        DBUE-Dropout  & \textbf{89.6} & 77.62 & 62.94 & 0.19 & \textbf{0.027} & \textbf{0.026} & \textbf{0.028} & \textbf{2.0}\\
        \bottomrule
    \end{tabular}%
    }
    \label{res tit}
\end{table}

\begin{figure}[h!]
    \centering
    \begin{subfigure}{\linewidth}
        \centering
        \includegraphics[width=1\linewidth]{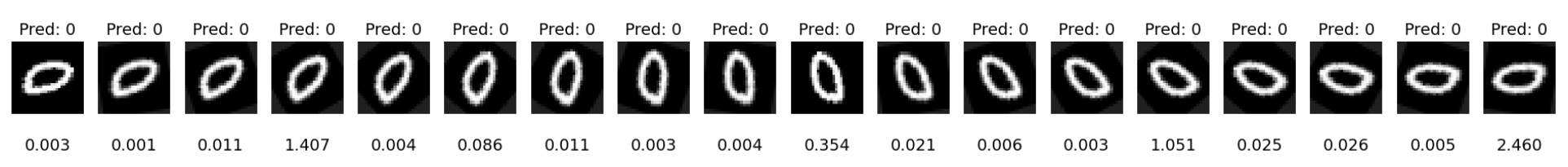}
    \end{subfigure}
    
    \begin{subfigure}{\linewidth}
        \centering
        \includegraphics[width=1\linewidth]{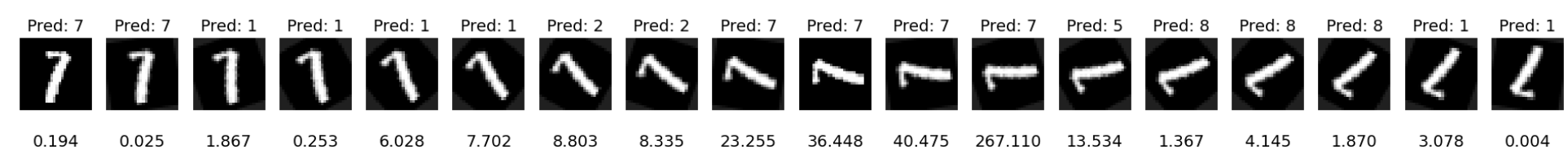}
    \end{subfigure}
    
    \begin{subfigure}{\linewidth}
        \centering
        \includegraphics[width=1\linewidth]{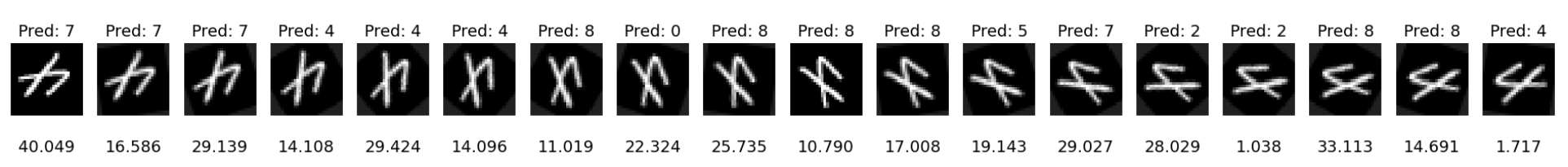}
    \end{subfigure}
    
    \caption{Rotated images with their associated uncertainties.}
    \label{rot_images}
\end{figure}
\noindent

\normalsize
\noindent
Table \ref{res tit} illustrates the classification performance as well as the UQ performance using the three methods on Titanic dataset. The uncertainty is calculated per sample and the median over train, test and noisy sets are given (to avoid effect of outliers).  It is noted that DEC and IAD are unable to detect OOD data since the uncertainty decreases while the accuracy is decreasing meaning the algorithm is confident in its wrong decisions. Meanwhile, DBUE-Dropout can efficiently detect the noisy and OOD data. 
 This means accuracy decreases on OOD data and uncertainty increases on noisy and OOD data.
 As for images, the DBUE-Dropout method performs well on the MNIST dataset. The method demonstrates high performance with a training accuracy of 99.95\%, a test accuracy of 99.25\%, a noisy test accuracy of 9.74\%, and an OOD accuracy of 6.78\%.
As for the uncertainty measure, it is 0.27 for training, 0.33 for testing, 0.53 for noisy inputs, and 19.19 for OOD data.
 The stranger the image becomes to the trained model, the higher the uncertainty. This can be shown by the increase in the median of uncertainty across different datasets. 
 
\noindent
To assess DBUE-Dropout's performance on images, some were rotated (0–360°), and their uncertainties were analyzed (Figure \ref{rot_images}). The digit 0 maintains low uncertainty due to its rotational invariance. As for the second image, uncertainty spikes at ambiguous angles, especially when horizontal, but remains low when clearly resembling a 1 or 7. 
The third row represents an image of digit 4 highlights that even correct predictions can exhibit high uncertainty. For instance, when inverted, uncertainty remains elevated due to visual ambiguity. However, as the digit returns to an upright position, uncertainty drops from almost 14 to 2, reinforcing the model’s confidence in more recognizable orientations.

\section{Conclusion}
A new method for uncertainty quantification in classification tasks is proposed.
The idea is to construct a deep evidential neural network with MC dropout. Given different samples of the class probability, the parameters of the Dirichlet distribution are calculated and used to evaluate the uncertainty.
Results confirm the overall performance in terms of classification and uncertainty quantification. Besides, comparison with state-of-the-art methods illustrates the potential of the proposed method to detect distribution shift through noisy and out-of-distribution (OOD) data across both tabular and image datasets. 
Future work will focus on enhancing the method to improve performance on noisy data and generalize to imbalanced datasets.

\bibliographystyle{unsrt}  
\bibliography{biblio}

\begin{thebibliography}{1}

\bibitem{Neal95}
Radford~M. Neal.
\newblock Bayesian learning for neural networks.
\newblock 1995.

\bibitem{lakshminarayanan17}
Balaji Lakshminarayanan, Alexander Pritzel, and Charles Blundell.
\newblock Simple and scalable predictive uncertainty estimation using deep
  ensembles.
\newblock {\em Advances in neural information processing systems}, 30, 2017.

\bibitem{malinin18}
Andrey Malinin and Mark Gales.
\newblock Predictive uncertainty estimation via prior networks.
\newblock {\em Advances in neural information processing systems}, 31, 2018.

\bibitem{Gal16}
Gal.Y and Ghahramani.Z.
\newblock Dropout as a bayesian approximation: Representing model uncertainty
  in deep learning.
\newblock 2016.

\bibitem{sensoy18}
Murat Sensoy, Lance Kaplan, and Melih Kandemir.
\newblock Evidential deep learning to quantify classification uncertainty.
\newblock {\em Advances in neural information processing systems}, 31, 2018.

\bibitem{tsiligkaridis21}
Theodoros Tsiligkaridis.
\newblock Information aware max-norm dirichlet networks for predictive
  uncertainty estimation.
\newblock {\em Neural Networks}, 135:105--114, 2021.

\bibitem{minka2000}
Thomas Minka.
\newblock Estimating a dirichlet distribution, 2000.

\bibitem{jsang18}
Audun Jsang.
\newblock {\em Subjective Logic: A formalism for reasoning under uncertainty}.
\newblock Springer Publishing Company, Incorporated, 2018.

\bibitem{loeve1977elementary}
Michel Lo{\`e}ve and M~Lo{\`e}ve.
\newblock {\em Elementary probability theory}.
\newblock Springer, 1977.

\end{thebibliography}

\end{document}